\pdfoutput=1

\documentclass[11pt]{article}

\usepackage[]{EMNLP2023}

\usepackage{times}
\usepackage{latexsym}

\usepackage[T1]{fontenc}

\usepackage[utf8]{inputenc}

\usepackage{microtype}

\usepackage{inconsolata}

\usepackage{amsmath,amssymb,bm,mathtools}
\usepackage{multirow}
\usepackage{booktabs} 
\usepackage{graphicx}
\usepackage{soulutf8}
\usepackage{subfig}
\usepackage{linguex}
\usepackage{tipa}
\title{Homophone Disambiguation Reveals Patterns of Context Mixing in\\Speech Transformers}

\author{
Hosein Mohebbi$^{1}$ ~ Grzegorz Chrupała$^{1}$ ~ Willem Zuidema$^{2}$ ~ Afra Alishahi$^{1}$ \\
$^1$ CSAI, Tilburg University ~
$^2$ ILLC, University of Amsterdam \\
\texttt{\{h.mohebbi, a.alishahi\}@tilburguniversity.edu}\\
\texttt{grzegorz@chrupala.me}\\
\texttt{w.h.zuidema@uva.nl}\\
}
  
\begin{document}
\maketitle

\begin{abstract}
  Transformers have become a key architecture in speech processing,
  but our understanding of how they build up representations of
  acoustic and linguistic structure is limited.  In this study, we
  address this gap by investigating how measures of `context-mixing'
  developed for text models can be adapted and applied to models of
  spoken language.  We identify a linguistic phenomenon that is ideal
  for such a case study: homophony in French (e.g.\ \emph{livre} vs
  \emph{livres}), where a speech recognition model has to attend to
  \emph{syntactic cues} such as determiners and pronouns in order to
  disambiguate spoken words with identical pronunciations and
  transcribe them while respecting grammatical agreement. We perform a series of controlled
  experiments and probing analyses on Transformer-based speech 
  models. Our findings reveal that representations in encoder-only
  models effectively incorporate these cues to identify the correct
  transcription, whereas encoders in encoder-decoder models mainly
  relegate the task of capturing contextual dependencies to decoder
  modules.\footnote{Code is freely available at \url{https://github.com/hmohebbi/ContextMixingASR}}
\end{abstract}

\section{Introduction}
\label{sec:introdiuction}
In both speech and text processing, variants of the Transformer
architecture \citep{NIPS2017_3f5ee243} have become ubiquitous. The key advantage of this neural
network topology lies in the modeling of pairwise relations between
elements of the input (tokens): the representation of a token at a
particular Transformer layer is a function of the weighted sum of the transformed
representations of all the tokens in the previous layer. This feature
of Transformers is known as \emph{context mixing} and understanding
how it functions in specific model layers is crucial for tracing the
overall information flow.  Several works
\citep[e.g.][]{kobayashi-etal-2020-attention,kobayashi-etal-2021-incorporating,mohebbi-etal-2023-quantifying}
have proposed methods to quantify context mixing and applied them to
transformers trained on text data. In the current study we investigate
context mixing in models trained on spoken language data.

We focus on the task of automatic speech recognition (ASR), which
consists of transcribing spoken utterances into their written
equivalent. In order to probe context mixing for speech models, we
exploit a quirk of the French spelling system: modern spoken French
does \textbf{not} in general\footnote{There is a small number of irregular
  exceptions to this phenomenon.} mark grammatical number on nouns, adjectives or
third-person verbs, only on determiners and pronouns -- however
written French reflects an earlier period of the language and \textbf{does}
mark nouns, adjectives and verbs for number. Therefore, for an ASR
system to produce grammatical transcriptions which feature number
agreement, the model needs to rely on very specific cues in the input
and/or in the prefix of the generated output.

For example, when transcribing the utterance \textit{Elle a perdu
  \underline{les} \textbf{livres}}, meaning ``She lost the books'',
when generating the bolded \textsc{target} word, the model needs to
decide whether to output \textit{livre} (singular) or rather
\textit{livres} (plural) and it cannot use the speech signal
corresponding to the target word to choose the correct form, as in spoken
French both are pronounced /liv\textinvscr/. The correct choice is determined by
grammatical agreement with the underlined \textsc{cue} word, here the plural
determiner, which is different from the singular form, both in speech
(/le/ vs /l\textschwa/) and in writing (\textit{les} vs \textit{le}).

Thanks to this phenomenon we can form a strongly supported prior
expectation that the representation of the cue token should
contribute to the representations of the target. We are therefore able
to validate previously suggested context mixing scoring
methods on a dataset of French utterances featuring such grammatical
homonyms: we find that context-mixing scoring methods, and especially Value Zeroing,
behave plausibly, largely agreeing with prior expectations, as well as with
targeted diagnostic probes and faithfulness tests.

We furthermore apply context-mixing scoring methods to two types of
architectures: encoder-decoder and encoder-only, and discover
interesting differences in how they behave. In encoder-only models,
spoken cue-word representations contribute to target representations
in a straightforward manner. Meanwhile, in encoder-decoder
architectures it is mostly the decoder which ensures grammatical
number agreement, and therefore spoken cue-word representations
contribute little to targets, while written cue-words in the generated
prefix make a decisive contribution.

\section{Related Work}
Our work builds on two strands of research: one focusing on analyzing models  of spoken language, and the other 
focusing specifically on quantifying information flow due to context-mixing in Transformer models of textual data.
\subsection{Analyses of speech models}
A large body of work on analyzing Transformer-based speech models has focused on probing their representations to uncover various types of linguistic structures they are hypothesized to encode, such as phonemic, lexical, syntactic information, as well as non-linguistic information such as speaker identity and demographics \citep{Pasad2021LayerWiseAO, Shah2021WhatAD, millet-dunbar-2022-self, Yang2023InvestigatingPA, shen23_interspeech}. \citet{Pasad2021LayerWiseAO} showed that in pre-trained wav2vec 2.0 models \citep{baevski2020wav2vec}, the representations evolve layer by layer according to an acoustic-linguistic hierarchy: the shallowest layers primarily encode acoustic features, followed by layers encoding phonetic, lexical, and semantic information. This trend then reverses in the deeper layers, unless the model is fine-tuned for ASR.
More recently, \citet{shen23_interspeech} showed that pre-trained wav2vec 2.0 models capture some degree of syntactic structure, particularly in the middle layers of the network. 

Another strand of research is focused on analyzing self-attention weights inside a Transformer layer as a simple measure of pairwise interaction of frame representations. By analyzing self-attention weights in Conformer \citep{Gulati2020ConformerCT}, \citet{shim2022understanding} observe two distinct patterns for phonetic and linguistic localization in the lower and upper layers, respectively. \citet{audhkhasi22_interspeech} measure the diversity of attention heads in Conformer and find that the attention probabilities from different heads show low diversity (analogous to the redundant pattern of attention in pre-trained language models) during the course of training, implying that they mostly focus on the same frames. In contrast, the value vectors show high diversity among different heads. 
\subsection{Analyses of context mixing}
While self-attention weights were initially deemed as the primary source of information mixing in Transformers \citep{clark-etal-2019-bert, kovaleva-etal-2019-revealing, reif2019visualizing, lin-etal-2019-open}, recent studies have shown that integrating other components in the architecture can yield more reliable measures \citep{kobayashi-etal-2020-attention, kobayashi-etal-2021-incorporating, modarressi-etal-2022-globenc, ferrando-etal-2022-measuring, mohebbi-etal-2023-quantifying}. \citet{kobayashi-etal-2020-attention} proposed a context-mixing technique called Attention Norm which is based on the norm of the transformed value vectors, and demonstrated that the model may assign a high attention weight to a token with a small norm, particularly for highly frequent and less informative words. More recently, \citet{mohebbi-etal-2023-quantifying} introduced a technique called Value Zeroing, which offers more reliable context mixing predictions compared to alternatives, mainly due to incorporating the output representations of Transformer layers that encompass all the components within the layer, including residual connections and feed-forward networks. 

These techniques are developed for and evaluated on text.  In this study we borrow the state-of-the-art context mixing methods, Attention Norms and Value Zeroing, and extend their applicability to both encoder-only and encoder-decoder-based speech models.

\section{Context Mixing}
Our goal is to quantify the contribution of a cue word to the representation of a target word across layers of speech Transformers. Here we first review the formation of contextualized representations in these models, and then describe how we adapt text-based context mixing methods for our purpose.
\subsection{Background}

\paragraph{Transformer Encoder.}
A speech encoder takes as input a sequence of audio representations known as audio \emph{tokens}, or \emph{frames}, and passes them through a stack of encoders to make contextualized audio representations. These tokens  are typically the output of a convolutional feature extractor module applied to the acoustic waveform.
Given a sequence of $d$-dimensional input audio representations $(\bm{x}_1, ..., \bm{x}_T)$, the goal of a Transformer encoder layer is to build contextualized audio representations $(\bm{\tilde{x}}_1, ..., \bm{\tilde{x}}_T)$ for each frame in the context.
This process consists of two sublayers: a multi-head self-attention mechanism (MHA) and a position-wise fully connected feed-forward network (FFN).

After passing through a layer normalization (LN)\footnote{As apposed to the original Transformer architecture \citep{NIPS2017_3f5ee243}, models studied in this paper utilize pre-LN rather than post-LN.}, each input vector in MHA module is compressed into separate query $(\bm{q}_i^h)$, key $(\bm{k}_i^h)$, and value $(\bm{v}_i^h)$ vectors via trainable linear transformations  for each head $h \in \{1,...,H\}$:
\begin{equation}
\label{eq:query}
{\bm{q}_i^h = \text{LN}_\text{MHA}(\bm{x}_i) \bm{W_Q^h} + \bm{b}_Q^h}
\end{equation}
\begin{equation}
\label{eq:key}
{\bm{k}_i^h = \text{LN}_\text{MHA}(\bm{x}_i) \bm{W_K^h} + \bm{b}_K^h}
\end{equation}
\vspace{-0.3cm}
\begin{equation}
\label{eq:value}
{\bm{v}_i^h = \text{LN}_\text{MHA}(\bm{x}_i) \bm{W_V^h} + \bm{b}_V^h}
\end{equation}
The raw attention weight $\alpha_{i,j}$, representing how much the $i^\text{th}$ frame attends to the $j^\text{th}$ frame within the context, is then computed by scaled dot-product between the corresponding query and key vectors followed by the softmax function:
\begin{equation}
\label{eq:raw_attentions}
\alpha_{i,j} = \mathop{\rm{softmax}} \left(\frac{\bm{q}_i \bm{k}_j^\top}{\sqrt{d_h}} \right) \in \mathbb{R}
\end{equation}
Consequently, the context vector output from MHA can be formulated as a weighted sum over the transformed value vectors within all frames across attention heads:
\begin{equation}
\label{eq:MHA}
{\bm z}_i = \sum^H_{h=1} \sum_{j=1}^T \mathbf{\alpha}_{i,j}^{h} \bm v_{j}^{h} \mathbf{W}_O^{h} + \bm b_{O} + \bm x_{i}
\end{equation}

Finally, two linear transformations with a GELU activation function \citep{Hendrycks2016GaussianEL} in between are applied to every ${\bm z}_i$ to produce output frame representation $\widetilde{\bm x}_i$:
\begin{equation}
\widetilde{\bm x}_i = {\rm{GELU}}\bigg(\text{LN}_\text{FFN}({\bm z}_i) \mathbf{W}_1 + \bm b_{1}\bigg) \mathbf{W}_2 + \bm b_{2} + \bm z_{i}
\end{equation}

\paragraph{Transformer Decoder.}
\label{sec:transformer_decoder}
In encoder-decoder-based ASR models, the model receives textual tokens as input to generate a probability distribution over the target vocabulary. The decoder  then constructs contextualized representations based on both the previously generated text tokens from earlier time steps and the final encoded audio representation.
The procedure is similar to encoders, except they perform the MHA procedure (Eq.~\ref{eq:MHA}) twice in two sublayers. The first MHA sublayer is modified to prevent tokens attending to subsequent positions when forming context vectors. The second MHA sublayer uses the key and value vectors from the last encoder layer to incorporate encoded audio representations into computation through cross-attention.
\paragraph{Decoder-less ASR.}
For the encoder-only architecture, ASR is carried out by outputting transcription tokens directly via a connectionist temporal classification (CTC) layer on top of the encoder output.
\subsection{Analysis methods}
\label{sec:methods}

In an encoder-decoder model, we compute three types of context mixing scores: within-encoder, within-decoder and cross (between encoder and decoder).
\begin{description}
\item[Within-encoder scores] measure the contribution of audio frame representations to a target frame representation in the encoder.
\item[Within-decoder scores] measure the contribution of token representations from previous time steps to a target token in the decoder.
\item[Cross scores] measure the contribution of audio frame representations from the encoder to a target token representation in the decoder.
\end{description}
To compute within-encoder and cross scores for audio input, we need to extract frames that correspond to each cue or target word (based on the alignment between the audio and the transcription). Consider sets $\mathcal{I}$ and $\mathcal{J}$, representing the frame indices for words $i$ and $j$, respectively. These sets allow us to aggregate word-level context mixing scores ($S_{i \leftarrow j}$) to measure how much the $j^\text{th}$ word in the context contributes to the representation of the $i^\text{th}$ word. 
In the decoder, these sets contain all the subword tokens corresponding the word of interest, and thus often contain just a single value.

We describe each context mixing method below.

\paragraph{Attention (Attn).} 
Raw attention weights are derived from Eq.~\ref{eq:raw_attentions}, averaged over all frames for each of the two words, over all heads: 
\begin{equation}
S_{i \leftarrow j} = \frac{1}{|\mathcal{I}| |\mathcal{J}| H} \sum_{n \in \mathcal{I}} \sum_{m \in \mathcal{J}} \sum^H_{h=1} \mathbf{\alpha}_{n,m}^{h}
\end{equation}

\paragraph{Attention Norm (AN).} Motivated by Eq.~\ref{eq:MHA}, this metric measures the Euclidean norm of the weighted transformed vector to measure each token’s contribution in a self-attention module \citep{kobayashi-etal-2020-attention}:
\begin{equation}
S_{i \leftarrow j} = \frac{1}{|\mathcal{I}| |\mathcal{J}| H} \sum_{n \in \mathcal{I}} \sum_{m \in \mathcal{J}} \sum^H_{h=1} \lVert \mathbf{\alpha}_{n,m}^{h} \bm v_{m}^{h} \mathbf{W}_O^{h} \rVert
\end{equation}

\paragraph{Value Zeroing (VZ).}
It measures how much the output representation of token $i$ is affected when excluding the $j^\text{th}$ token 
by zeroing its value vector \citep{mohebbi-etal-2023-quantifying}. In the speech context, we set the value vectors in all frames in $\mathcal{J}$ (corresponding to word $j$) to zero, extract the alternative representations for each frame $n\in\mathcal{I}$ (that is, $\bm{\tilde{x}}_n^{\neg j}$), and measure how much each frame representation has changed compared to the original ones:
\begin{equation}
\label{eq:VZ}
    S_{i \leftarrow j} = \frac{1}{|\mathcal{I}|} \sum_{n \in \mathcal{I}}\text{cos} (\bm{\tilde{x}}_n, \bm{\tilde{x}}_n^{\neg j})
\end{equation}

\section{Experimental Setup}
\subsection{Data}
As a case study for analyzing word-level context mixing in audio representations, we opt for homophony in French, where certain grammatical forms of a word are identical in pronunciation, despite having different spellings. The spoken forms are thus disambiguated in writing by relying on grammatical agreement phenomena, where nouns agree in number with determiners, and verbs agree with subject pronouns. The nature of this task makes it suitable for analyzing context mixing in speech models since it offers a strong hypothesis regarding which part of the context is most relevant when transcribing ambiguous target words. 
\begin{table*}[th!]
\centering
\resizebox{0.95\linewidth}{!}{
\begin{tabular}{llc}
\toprule
{\bf Pattern} & {\bf Examples of transcription} & {\bf \#} \\ 
\midrule
\multirow{2}{*}{Det\_Noun} & C'est \ul{le} septième {\bf titre} de champion de Syrie de l'histoire du club & \multirow{2}{*}{720} \\
          & Il y mène \ul{une} {\bf vie} d’études et de recherches & \\
\midrule
\multirow{2}{*}{Pronoun\_Verb} & Chaque jour, leurs concurrents les voient sortir de pistes dont \ul{ils} {\bf ignorent} l'existence & \multirow{2}{*}{257} \\
            & \ul{On} y {\bf trouve} une plage naturiste  & \\
\midrule
\multirow{2}{*}{Det\_Noun\_Verb} & Peu après cette élimination, \ul{le} \ul{\bf club} et Alexander se {\bf séparent} à l'amiable & \multirow{2}{*}{~23} \\
                & À la fin, \ul{les} \ul{\bf enfants} se {\bf révoltent} et détruisent l’école. & \\
\bottomrule
\end{tabular}
}
\caption{Examples of the extracted audios from the Common Voice corpus based on defined patterns. Last column shows the number of examples obtained. Cue and Target words are \ul{underlined} and {\bf bolded}, respectively.} 
\label{Tab:data}
\end{table*}

We define three specific syntactic templates in which homophony may appear. All rely on grammatical number agreement: the first one captures determiner-noun number agreement; the second -- number agreement between subject pronoun and verb; and the third -- number agreement between subject noun phrase and verb. Table~\ref{Tab:data} shows examples of utterances captured using these templates.

We used the test set of the Common Voice \citep{ardila-etal-2020-common} spoken corpus\footnote{We only used the test split of the corpus since the training and validation sets have already been utilized during the fine-tuning phase of models within the Wav2vec2 family.}, which consists of 26 hours of transcribed speech in French, and generated dependency trees of transcriptions with SpaCy \citep{spacy2} to discover instances of these templates. We selected those examples for which all models in our analysis predict the cue and target words correctly.
This resulted in the extraction of 1000 audio-transcription pairs.\footnote{The details of the data extraction procedure is described in Appendix~\ref{App:data_extraction}.}

\subsection{Target models}
Our experiments cover five prominent off-the-shelf models based on Transformers for learning speech representations, which can be classified into two categories in terms of their architecture and training objectives. See Table~\ref{Tab:models} for an overview.

\begin{table*}[th!]
\centering
{\small
\resizebox{0.95\linewidth}{!}{
\begin{tabular}{lccccrcc}
\toprule
Model & Architecture & Size & Layers & Width & Pre-training & Multilingual & Fine-tuned\\ 
\midrule 
\multirow{3}{*}{Whisper} & \multirow{3}{*}{encoder-decoder} & base & 6 & 512 & \multirow{3}{*}{\text{supervised}} & \multirow{3}{*}{yes} & \multirow{3}{*}{no} \\
 & & small & 12 & 768 & & & \\
 & & medium & 24 & 1024 & & & \\
\midrule
XLSR-53 & \multirow{2}{*}{encoder} & \multirow{2}{*}{large} & \multirow{2}{*}{24} & \multirow{2}{*}{1024} & \multirow{2}{*}{\text{self-supervised}} & yes & yes  \\
XLSR-1 & & & & & & no & yes\\
\bottomrule
\end{tabular}
}
\caption{Specification of the target models in our study.} 
\label{Tab:models}
}
\end{table*}

\paragraph{Whisper family.}
Whisper \citep{Radford2022RobustSR} is a multilingual encoder-decoder multitask speech model. It takes raw audio as input, splits it into 30-second chunks, and transforms them into log-Mel spectrograms. After passing through two convolution layers, these audio features are then fed into a stack of Transformer encoder layers to create contextualized audio representations. The model is autoregressively trained to predict next token on a set of supervised audio-to-text tasks such as language identification, phrase-level timestamps, multilingual speech transcription, and speech translation to English. For our experiments, we used Whisper in three different sizes (base, small, medium).

\paragraph{Wav2vec2 family.}
This family of models employs only the encoder part of the Transformer architecture, built on the wav2vec~2.0 framework \citep{baevski2020wav2vec}. The model receives raw  audio waveform input and maps it into latent speech representations via a multi-layer convolutional neural network. These latent representations are partially masked \cite{Jiang2019ImprovingTS, Wang2020UnsupervisedPO} and passed through a stack of Transformer encoder layers to build contextualized representations. The model is then trained by solving a contrastive task over quantized speech representations in a self-supervised manner. After pre-training, the model can be fine-tuned on labeled data using a Connectionist Temporal Classification (CTC) loss \citep{10.1145/1143844.1143891} for downstream speech recognition. 
For our experiments, we used XLSR-53\footnote{\url{https://huggingface.co/jonatasgrosman/wav2vec2-large-xlsr-53-french}} \citep{Conneau2020UnsupervisedCR}, and XLSR-1\footnote{\url{https://huggingface.co/LeBenchmark/wav2vec2-FR-7K-large}} \citep{Evain2021LeBenchmarkAR} from this family. 
The former is pre-trained on 56K hours of speech data in 53 languages, whereas the latter is pre-trained on 7K hours of French speech only.
Both models are fine-tuned for French ASR using CTC.

\subsection{Frame-word alignment}
To investigate context mixing in speech models at the word level, we need to know which frames in the encoder representation correspond to which words in the audio. As the Common Voice dataset does not provide this information, we used Montreal Forced Aligner \citep[MFA]{mcauliffe17_interspeech} to extract the start ($t_s$) and end time ($t_e$) of each word in an utterance, and mapped them to boundary frames $f_s$ and $f_e$:
\begin{equation}
\label{eq:frame_word_alignment}
{f = \lceil \frac{t}{\mathcal{T}} \times T} \rceil
\end{equation}
where $\mathcal{T}$ and $T$ denote the total time of a given utterance and the total number of frames in the encoded audio, respectively.\footnote{In contrast to the models in the Wave2Vec 2.0 family, Whisper does not support attention mask in its encoder and instead pads the audio input with silence to a constant duration of 30 seconds in its processor. Therefore the variables $\mathcal{T}$ and $T$ in Eq.~\ref{eq:frame_word_alignment} are always set as 30 seconds and 1500, respectively for Whisper models.} The range ($f_s$, $f_e$) indicates which frames in the encoder correspond to the selected word. In our dataset, on average a word corresponds to 369ms of audio duration and 19 encoder frames.

\section{Experiments}
We design a number of experiments meant to first validate the context-mixing scoring methods in the context of speech transformers, and further to provide converging evidence on the specific patterns of context-mixing in the two target architectures on our dataset.
\subsection{Cue contribution}
\label{sec:cue-contribution}
To quantify the contribution of the cue word to the representation of a target word, we compute context mixing maps based on different methods described in Sec.~\ref{sec:methods}. For each method, we normalize the scores for all words in a sentence so that they are positive values and sum up to one. We then select the $t^\text{th}$ row of each context mixing map, where $t$ represents the time step at which a target word is being generated. The resulting 1-D array indicates how much the target word relies on other words in the context when forming its contextualized representation.

Recall that all models in our experiments accurately transcribe both cue and target words, as we specifically extracted the dataset under this condition. As cue words are the sole indicators in the context for identifying the correct transcription of target words, we expect that accurate disambiguation of the target word is due to the model's reliance on the cue words when constructing target word representations. To measure this reliance, we define a binary cue vector $C$ according to the following condition:
\begin{equation}
C_i = 
\begin{cases}
1, & \textrm{the $i^\text{th}$ position} \in \textrm{Cue word} \\
0, & \textrm{otherwise} \
\end{cases}
\label{eq:cue_vector}
\end{equation}
We then compute the {\it Cue contribution} score as the dot product of the cue vector ($C$) and a given context mixing score ($S_t$).

\begin{figure}[th!]
\centering
    \includegraphics[width=\columnwidth]{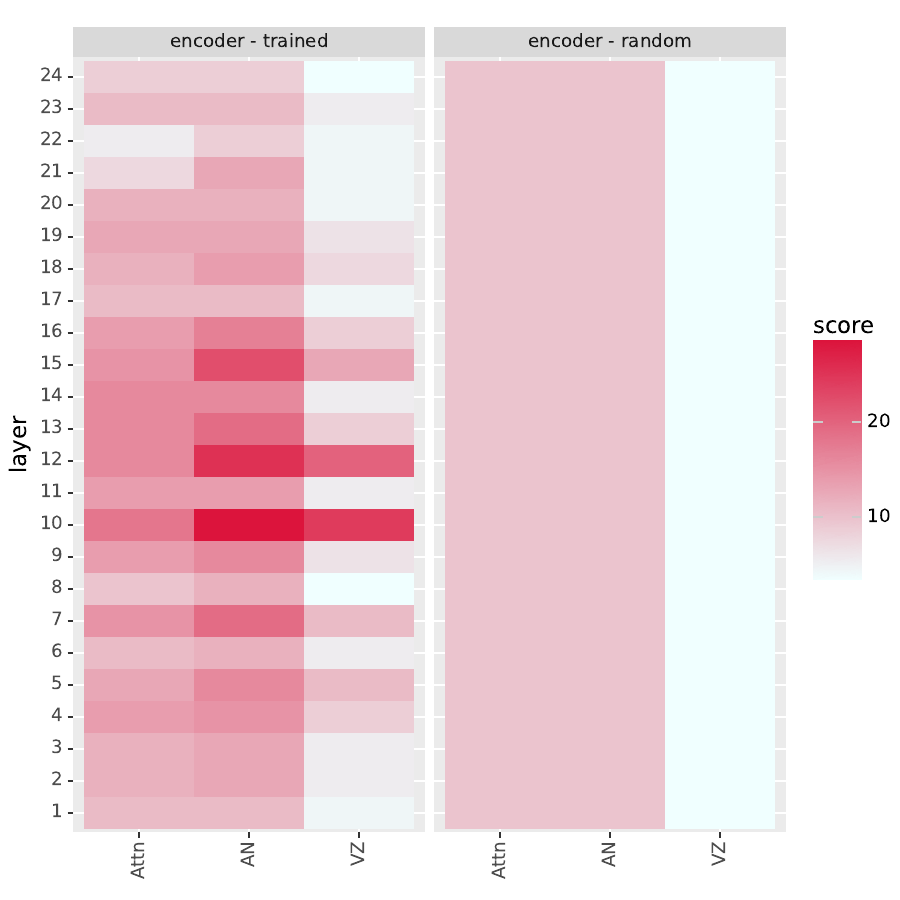}
    \caption{Layer-wise cue contribution according to different analysis methods averaged over all examples for XLSR-53, trained (left) vs.\ randomly initialized (right).} 
    \label{fig:cue_alignment_xlsr-53_all}
\end{figure}

\paragraph{Results.}
Figure~\ref{fig:cue_alignment_xlsr-53_all} shows cue contribution scores for all layers of both randomly initialized and trained versions of \mbox{XLSR-53}, which is an encoder-based model. For the model with random weights, the cue contribution scores are low and uniform across the layers. For the trained model, cue contribution scores indicate that cues are prominently integrated into the target audio representations in the middle layers (notably, at layers 10, 12 and 15). This pattern conforms to the prior expectations for the grammatical number homonymy task.

\begin{figure}[th!]
\centering
    \includegraphics[width=\columnwidth]{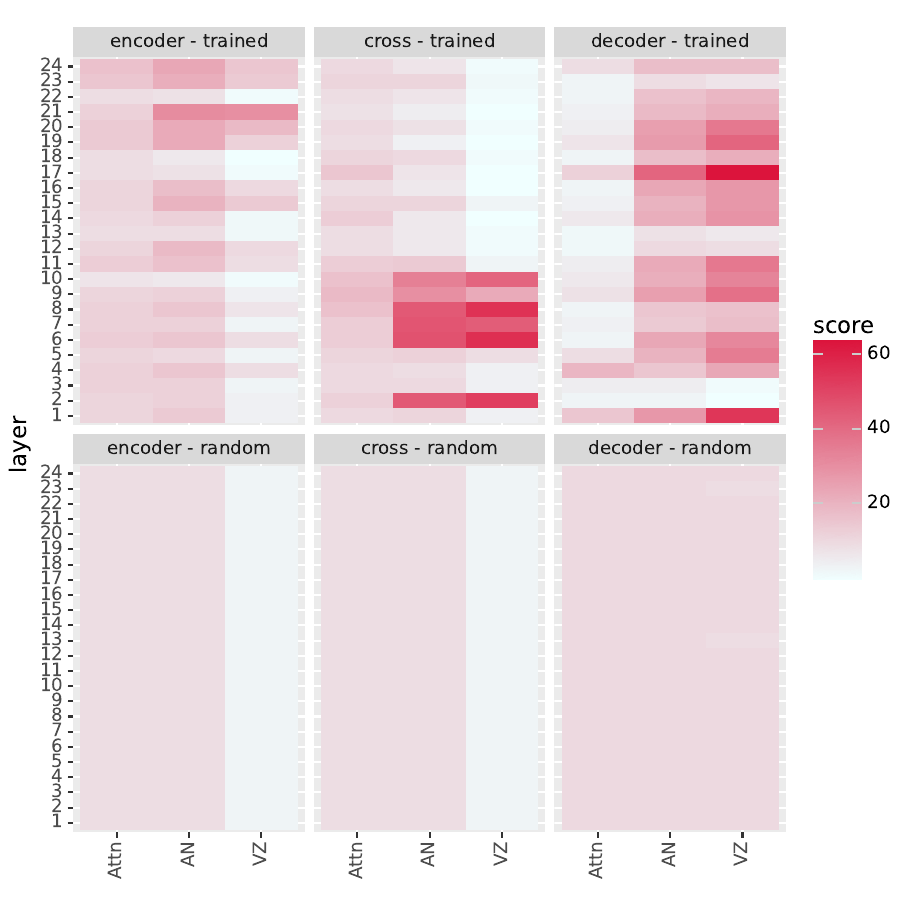}
    \caption{Layer-wise cue contribution according to different analysis methods averaged over all examples for Whisper-medium, trained (top) vs.\ randomly initialized (bottom).} 
    \label{fig:cue_alignment_whisper_medium_all}
\end{figure}

Figure~\ref{fig:cue_alignment_whisper_medium_all} displays the cue contribution scores for the Whisper-medium model. Results show that the within-encoder scores are not noticeably higher than the ones for the randomly initialized model, except for layers 19--21 and 23--24 which show a slight increase. In contrast, the cross and within-decoder scores show that the decoder prominently incorporates the cue word into its contextualized representation of the target word.

In the decoder, each layer has two options for information integration (cf.\ Eq.~\ref{eq:MHA}): attending to previously generated tokens or attending to the audio representations output from the last encoder layer. 
By comparing the cross and within-decoder alignment scores, we find that the representations in the decoder strongly capture the syntactic dependency on the cue word through the self-attention module. 
 We conjecture that in the encoder-decoder model, the decoder plays a crucial role in capturing syntactic dependency and does not rely on cue word representations in the encoder, probably due to having access to text-token representations as an easier alternative.
We examine the role of the cross-attention module more carefully in Section~\ref{sec:silencing}. 

For all these scores, predictions made by Value Zeroing and Attention Norms methods are largely consistent with each other, with Value Zeroing suggesting a stronger reliance on cue word in the decoder of the trained model. In contrast, raw attention scores are uniform and less informative, as shown also in previous studies \citep{kobayashi-etal-2020-attention, hassid-etal-2022-much, mohebbi-etal-2023-quantifying}.

\begin{figure}[ht!]
    \centering
    \includegraphics[width=\columnwidth]{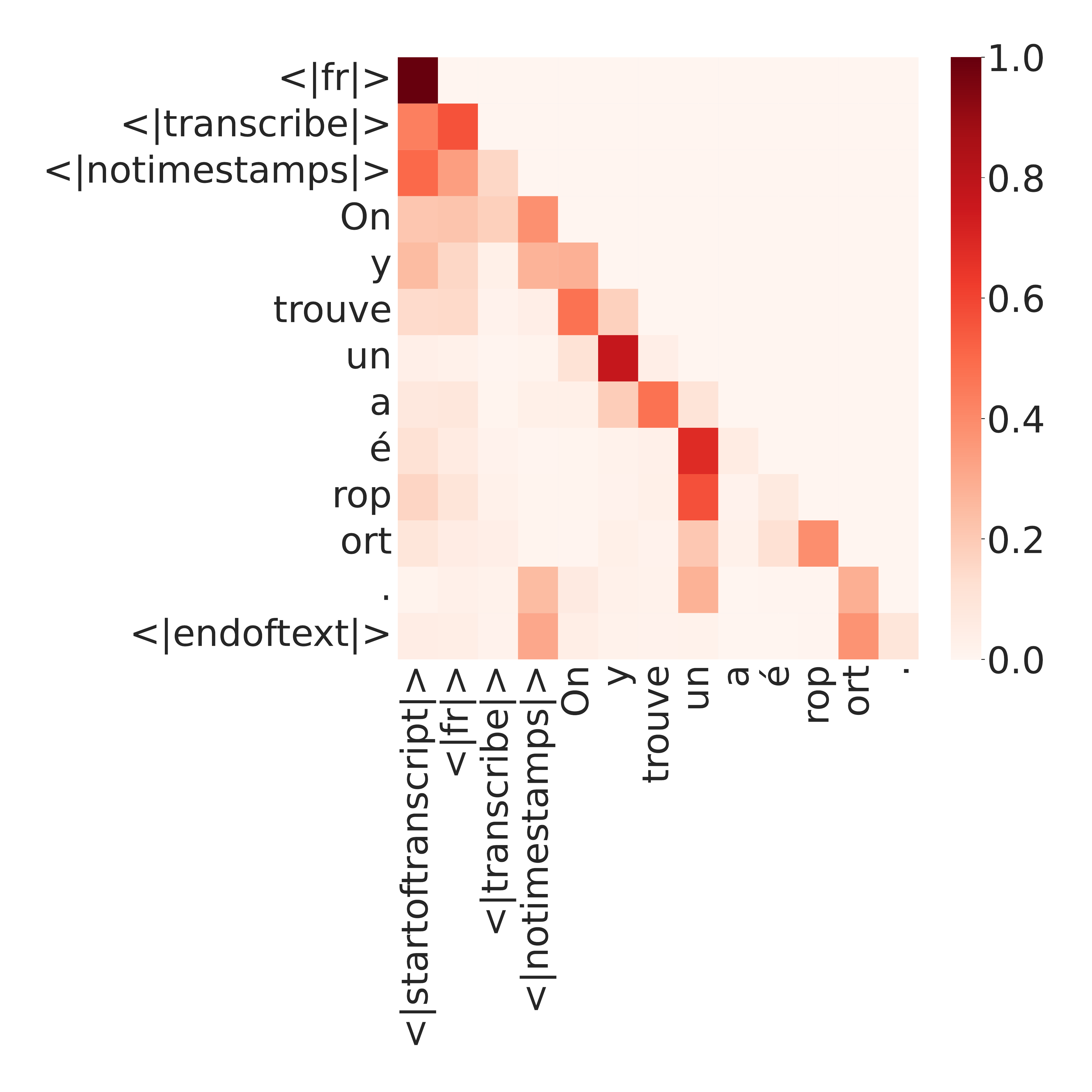}
    \caption{{\bf Within-decoder} Value Zeroing context mixing map for Whisper-medium at layer 10 for our running example. Target word `trouve' representations notably rely on the cue word `On' in the context. Similarly, the cue word `un' notably contributes to forming representations of the word `aéroport'.}
    \label{fig:examplewise_dec_L10}
\end{figure}

Figure~\ref{fig:examplewise_dec_L10} shows the context-mixing map for layer~10 for an example from the dataset: ``\emph{\ul{On} y {\bf trouve} un aéroport.}'', illustrating the dependence of target subword tokens on the cue word in the decoder prefix.


\subsection{Probing number encoding}
\label{sec:probing}
Since the main source of information for disambiguating the target word is its number agreement with the cue word, we hypothesize that layers with a higher cue contribution score must encode number information more strongly and/or accessibly.
To test our hypothesis, we conduct a probing experiment by extracting target word representations across all layers, and associating each representation with a \emph{Singular} or \emph{Plural} label. We then train a Logistic Regression classifier with L2 regularization on these representations to predict target grammatical number. 

\begin{figure}[t]
\centering
    \includegraphics[width=\columnwidth]{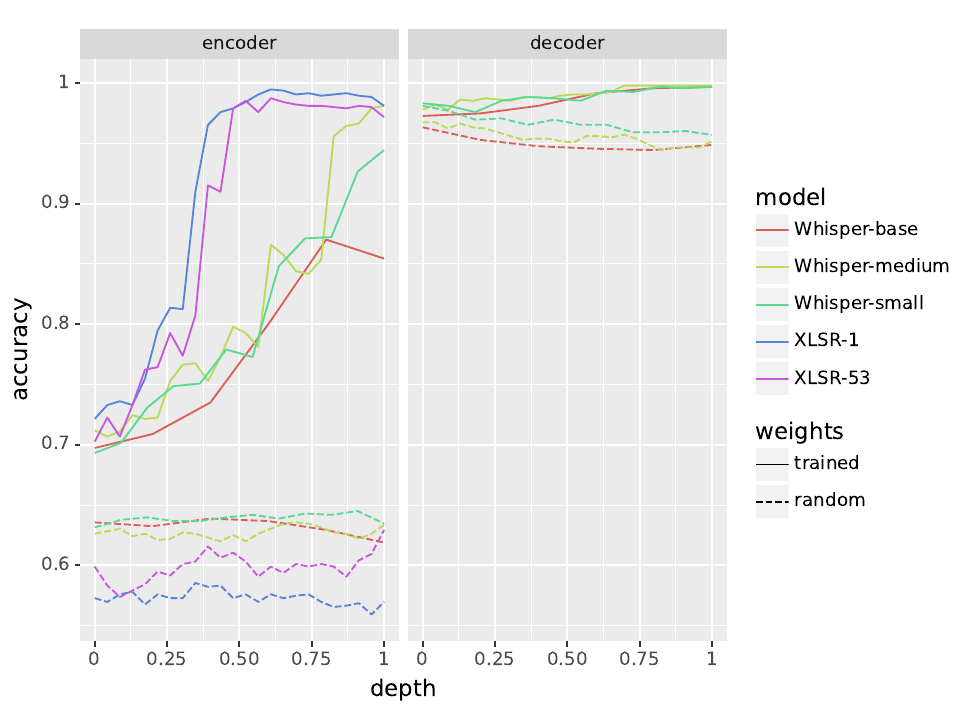}
    \caption{Accuracy of probing classifiers trained on frozen target representations obtained from various ASR models. The depth of Whisper-base (6) and Whisper-small (12),  has been normalized to 1 to facilitate comparisons.} 
    \label{fig:probing}
\end{figure}

\paragraph{Results.}
Figure~\ref{fig:probing} presents the 3-fold cross-validation accuracy of the probing classifiers for different models across all layers. As a baseline, 
we also include the results for target representations obtained from randomly initialized models.

For Whisper decoders, probing accuracy is close to one -- indeed the accuracy is very high even for randomly initialized target models suggesting that simple word identity is being used by the diagnostic classifier to predict grammatical number.

Encoder representations in encoder-based and encoder-decoder-based models, however, display two distinct patterns. Whisper encoders show a gradual increase as we progress through the layers. The encoder of Whisper-medium shows a jump at layer 20, the same layer for which we observed a significant cue contribution score in Section~\ref{sec:cue-contribution}. Comparing Whisper models with different sizes, we can see that the gap between encoder and decoder decreases as the model size increases. 
In contrast, XLSR-53 and XLSR-1 display a noticeable spike in the middle layers (notably at layer 10), the same layers for which we observed higher cue contribution scores. This implies that the target words in the middle layers of these models significantly incorporate the cue words in their representations and encode number. This information can be transferred to the final layer with minimal loss, where the model is able to correctly disambiguate the target word.

\subsection{Silencing the cue words}
\label{sec:silencing}
Results of the previous experiments in Section~\ref{sec:probing} suggest that in encoder-decoder models, target representations in the decoder show a stronger encoding of syntactic dependency on the cue word compared to encoder representations.
Additionally, our analysis of the context mixing scores in Section~\ref{sec:cue-contribution} revealed that encoder-decoder models exploit the agreement information provided by the cue word via its text representation in the decoder, rather than relying on the encoded spoken cue representation via the cross-attention module. 

In this section, we employ the concept of input ablation \citep{Covert2020ExplainingBR} to assess the influence of cue words on an ASR model's decision:
we measure how much the model's confidence drops when the cue word is excluded from the input during the transcription of ambiguous target words. For encoder models, we silence the frames corresponding to the cue word in the raw audio and provide the modified audio input to the model. For encoder-decoder models, we consider the following conditions: silencing the cue words from the raw audio input (silence cue), replacing the cue token with an unknown token in the input of the decoder (blank cue), and combining both modifications simultaneously (silence \& blank cue). As a control we also include the condition where we silence the target word in the input to the encoder (silence target) -- in this case we expect the models' confidence to drop irrespective of architecture, as this manipulation removes most of the information available to the model about the identity of the target word.

We compute the drop in model's output probability as $p({y_t} | \bm{e}) - p({y_t} | \bm{e} \backslash e_t)$, where $y_t$ denotes the logit value corresponding to the target word at time step $t$, and $\bm{e}$ refers to a given raw input audio at encoder setup, and original previously generated tokens at decoder setup.

\begin{figure}[t]
\centering
    \includegraphics[width=\columnwidth]{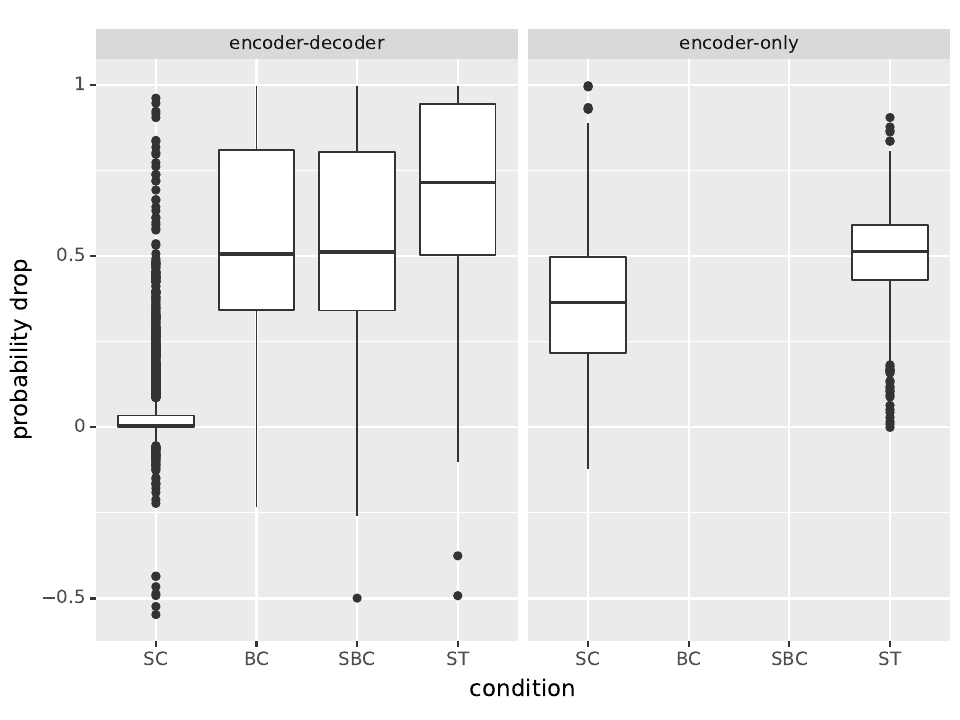}
    \caption{Changes in the models' confidence in generating the target word in different setups of input ablations, averaged over models within each group. SC=Silence cue, BC=Blank cue, SBC=Silence \& blank cue, ST=Silence target.} 
    \label{fig:silencing}
\end{figure}

\paragraph{Results.}
Figure~\ref{fig:silencing} presents the results for different ASR models, averaged over each model family. Notably, in encoder-decoder models (models in the Whisper family), we can see that silencing the cue word in the encoder input does not significantly affect the model's confidence when generating the target word. However, if we blank out the cue token in the decoder input while keeping the encoder input intact, the model experiences a significant decrease in confidence in its target word prediction. The same drop can also be observed when both the encoder and decoder inputs have the cue words silenced/blanked out. These findings support our hypothesis that in encoder-decoder models, the decoder captures the syntactic dependency on the cue words in the target representations by attending to the cue word from its prefix rather than relying on the encoded cue words from the encoder. 
The scenario for encoder-based models, however, is different. As we can see, silencing the cue words in the encoder input leads to a significant drop in the model's confidence in its target word prediction. This corroborates with the results observed in Sections~\ref{sec:cue-contribution} and \ref{sec:probing}.

\section{Conclusion}
In this paper, we adapted state-of-the-art context mixing methods from the text domain to analyze sensitivity to syntactic cues and to trace information mixing in Transformer-based models of speech recognition. Our analyses revealed that in encoder-only models, spoken cue-word representations directly contribute to the formation of target word representations. In contrast, encoder-decoder models largely relegate the task of capturing contextual dependencies to the decoder module. Through a series of complementary experiments, we identified two distinct roles performed by the decoder in encoder-decoder speech models when forming contextualized representations: attending to discrete prefix cues using self-attention, and attending to encoded audio from the encoder using cross-attention. 

Our study validates the applicability of the adapted context mixing scores that were originally developed for text, and showcases their potential for discovering patterns of information flow in Transformer models of language in spoken modality. 
We hope this convergence in analytical methodologies between speech and text communities can pave the way for developing unified models for natural language processing \cite{chrupala-2023-putting}.
In line with previous research, our results discredit raw attention weights as a reliable source of information for information mixing in speech models.

\section{Limitations}
Our experiments rely on a specific phenomenon in  French, as it provides a discrepancy between marking of number agreement in spoken versus written forms, and therefore provide us with an ideal case study for investigating context dependency and incorporating syntactic cues. However, in the case of homophony in French, there is one single cue word for each target word representation. In the future, we must search for other phenomena that allow us to investigate the role of multiple (converging or contrasting) sources of information when disambiguation is needed. 

When choosing our target models, we have focused on models that are trained for automatic speech recognition. Models trained according to different objectives must also be considered, and the impact of pre-training vs.\ fine-tuning on context dependency and information mixing must be examined. 

It is also worthwhile to use our adapted context-mixing methods to explore unsuspected phenomena in speech Transformers, without relying on strong prior hypotheses about model behavior.

\section*{Acknowledgments}
This publication is part of the project \emph{InDeep: Interpreting Deep Learning Models for Text and Sound} (with project number NWA.1292.19.399) of National Research Agenda (NWA-ORC) programme. Funding by the Dutch Research Council (NWO) is gratefully acknowledged. We also thank SURF (\url{www.surf.nl}) for the support in using the National Supercomputer Snellius.

\bibliography{custom}
\bibliographystyle{acl_natbib}

\appendix
\counterwithin{figure}{section}
\counterwithin{table}{section}

\section{Number homophony extraction procedure}
\label{App:data_extraction}
Besides matching our defined templates based on dependency trees, we considered a supplemental set of conditions to ensure homophony exists in our constructed data. 
\begin{itemize}
    \item Irregular nouns in the list [oeil,   yeux, aïeul, aïeux, ciel, cieux, vieil, vieux] as well as nouns that end with \emph{al} or \emph{ail} are  filtered out as target noun. 
    \item Target verbs must be in the present or imperfect tense, in any person-number except for 1-plural and 2-plural.
    \item Target verbs in their different number, person, and tense must end with specific letters to ensure they are pronounced the same in their different forms. In the present tense, a verb must end with \emph{e}, \emph{es}, \emph{e}, and \emph{ent} in their 1-singular, 2-singular, 3-singular, and 3-plural, respectively, while in the imperfect tense, the verb must end with \emph{ais}, \emph{ais}, \emph{ait}, and \emph{aient} in their 1-singular, 2-singular, 3-singular, and 3-plural, respectively. To check for this condition we utilized inflecteur library\footnote{\url{https://github.com/Achuttarsing/inflecteur}} which is based on the DELA dictionary\footnote{\url{https://infolingu.univ-mlv.fr}}.
\end{itemize}

\begin{figure}[h!]
\centering
    \includegraphics[width=.7\columnwidth]{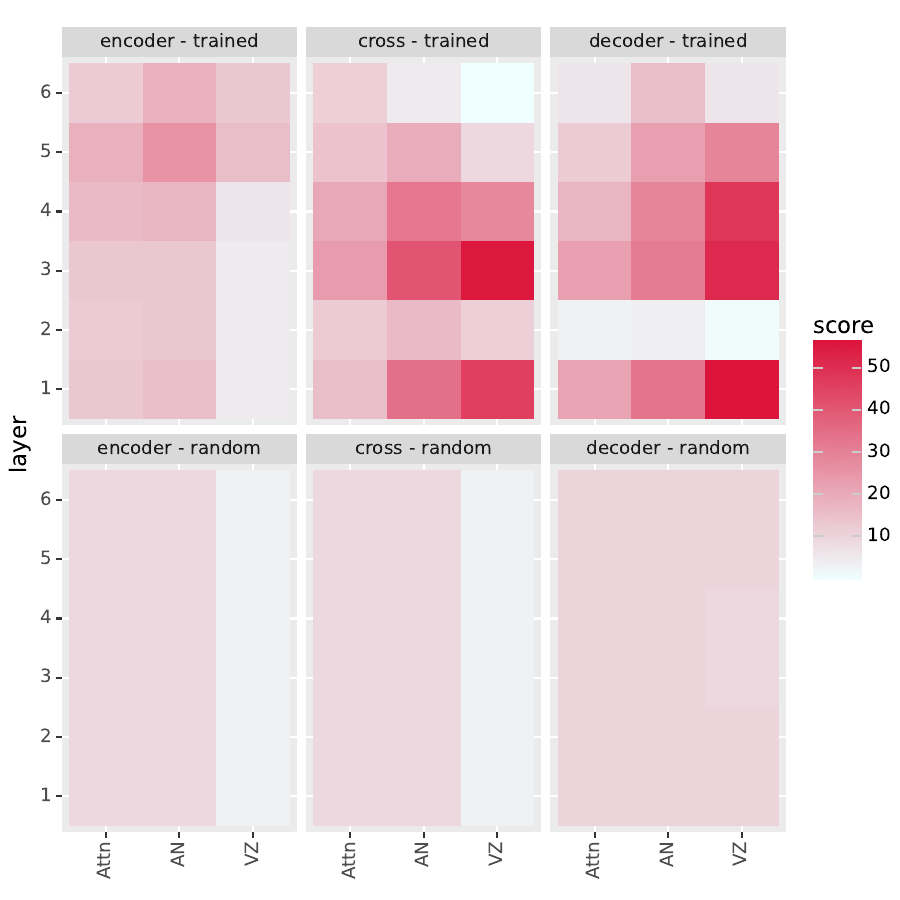}
    \caption{Layer-wise cue contribution according to different analysis methods averaged over all examples for \textbf{Whisper-base}, trained (top) vs.\ randomly initialized (bottom).} 
    \label{fig:cue_alignment_whisper_base_all}
\end{figure}

\section{Cue contribution for other model sizes}
The pattern of the results for layer-wise cue contribution is the same among all model sizes and other models within each family.
\label{app:cue_alignment_other_sizes}

\begin{figure}[th!]
\centering
    \includegraphics[width=\columnwidth]{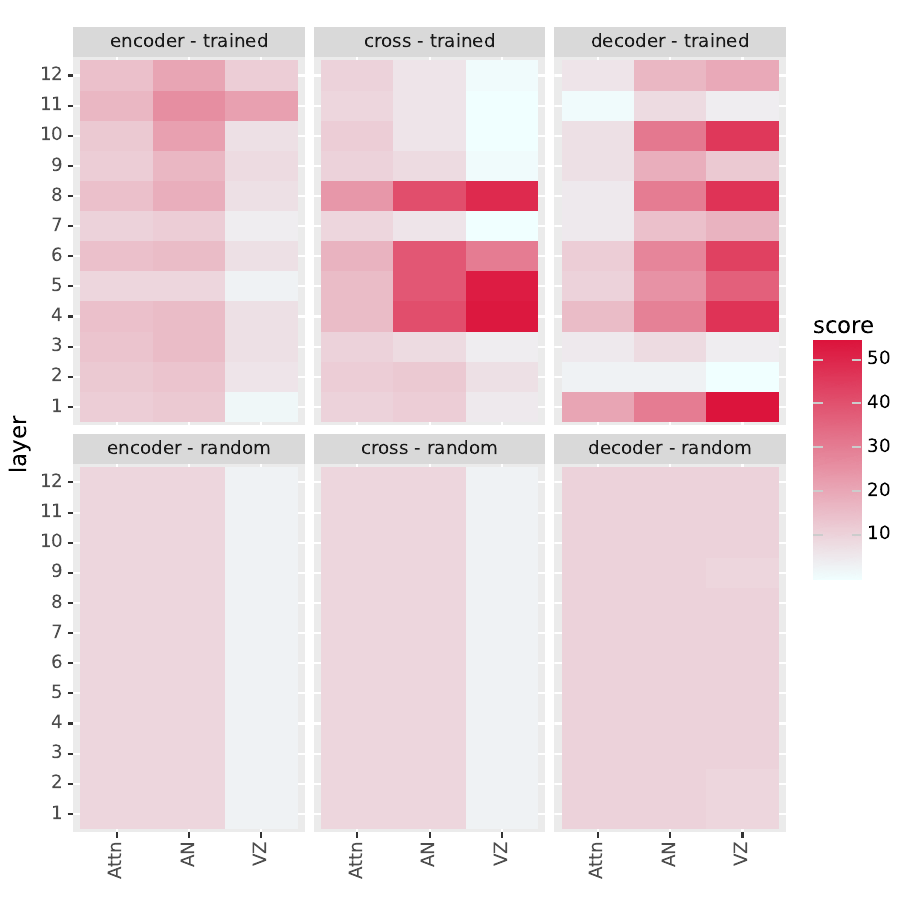}
    \caption{Layer-wise cue contribution according to different analysis methods averaged over all examples for \textbf{Whisper-small}, trained (top) vs.\ randomly initialized (bottom).} 
    \label{fig:cue_alignment_whisper_small_all}
\end{figure}

\begin{figure}[ht!]
\centering
    \includegraphics[width=\columnwidth]{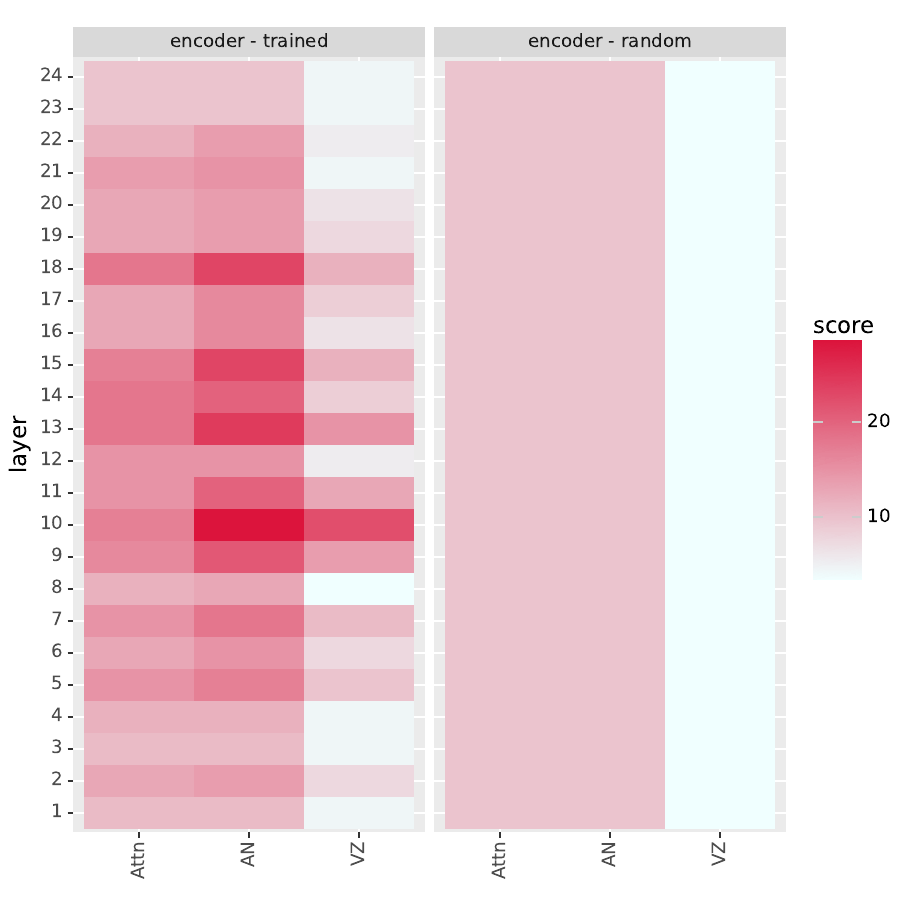}
    \caption{Layer-wise cue contribution according to different analysis methods averaged over all examples for \textbf{XLSR-1}, trained (left) vs.\ randomly initialized (right).} 
    \label{fig:cue_alignment_xlsr-1_all}
\end{figure}

\clearpage
\section{Example-wise visualization}
\begin{figure}[ht!]
    \centering
    \includegraphics[width=\columnwidth]{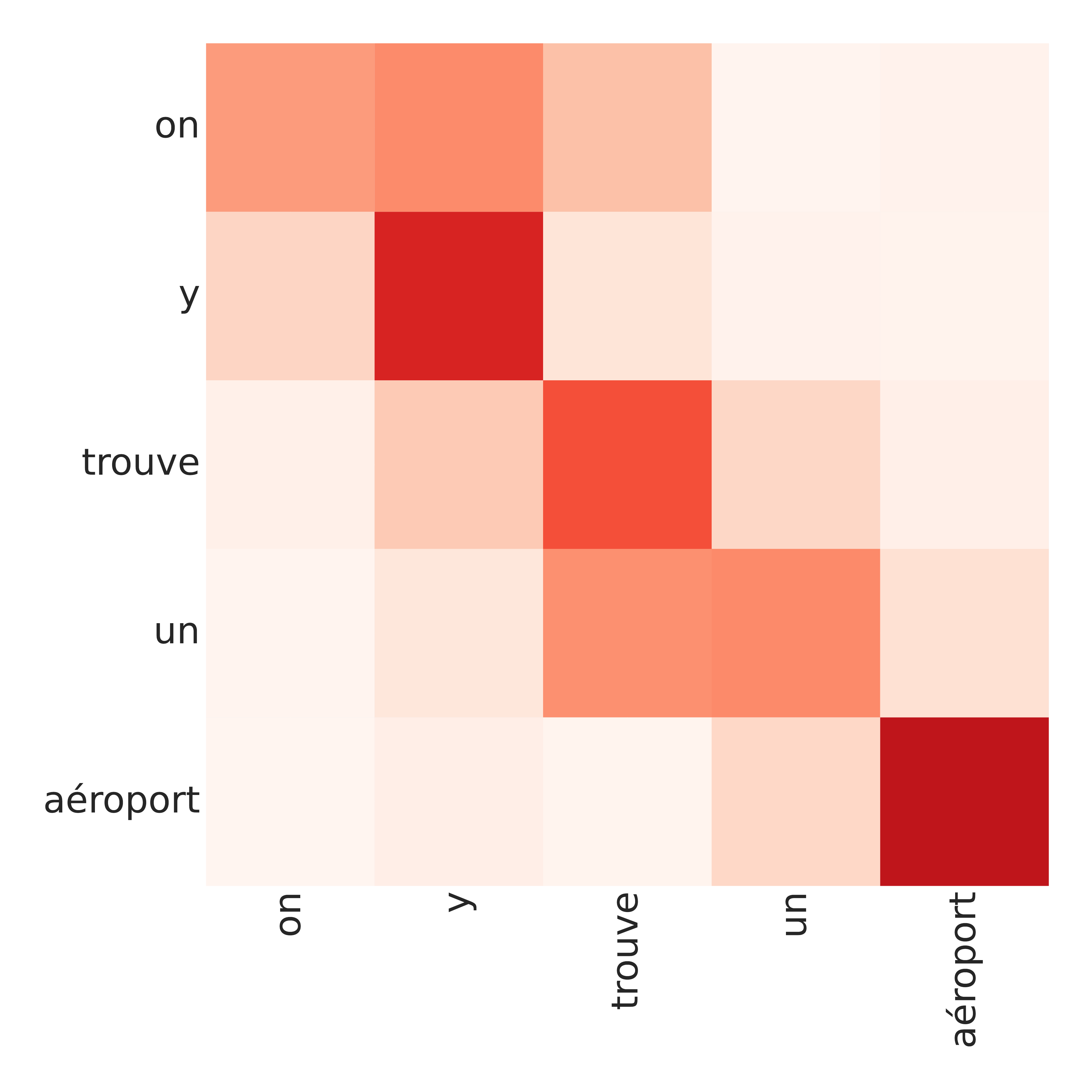}
    \caption{{\bf Within-encoder} Value Zeroing context mixing map for {\bf XLSR-53} at layer 10 for our running example.}
    \label{fig:examplewise_xlsr-53_enc_L10}
\end{figure}

\begin{figure}[ht!]
    \centering
    \includegraphics[width=\columnwidth]{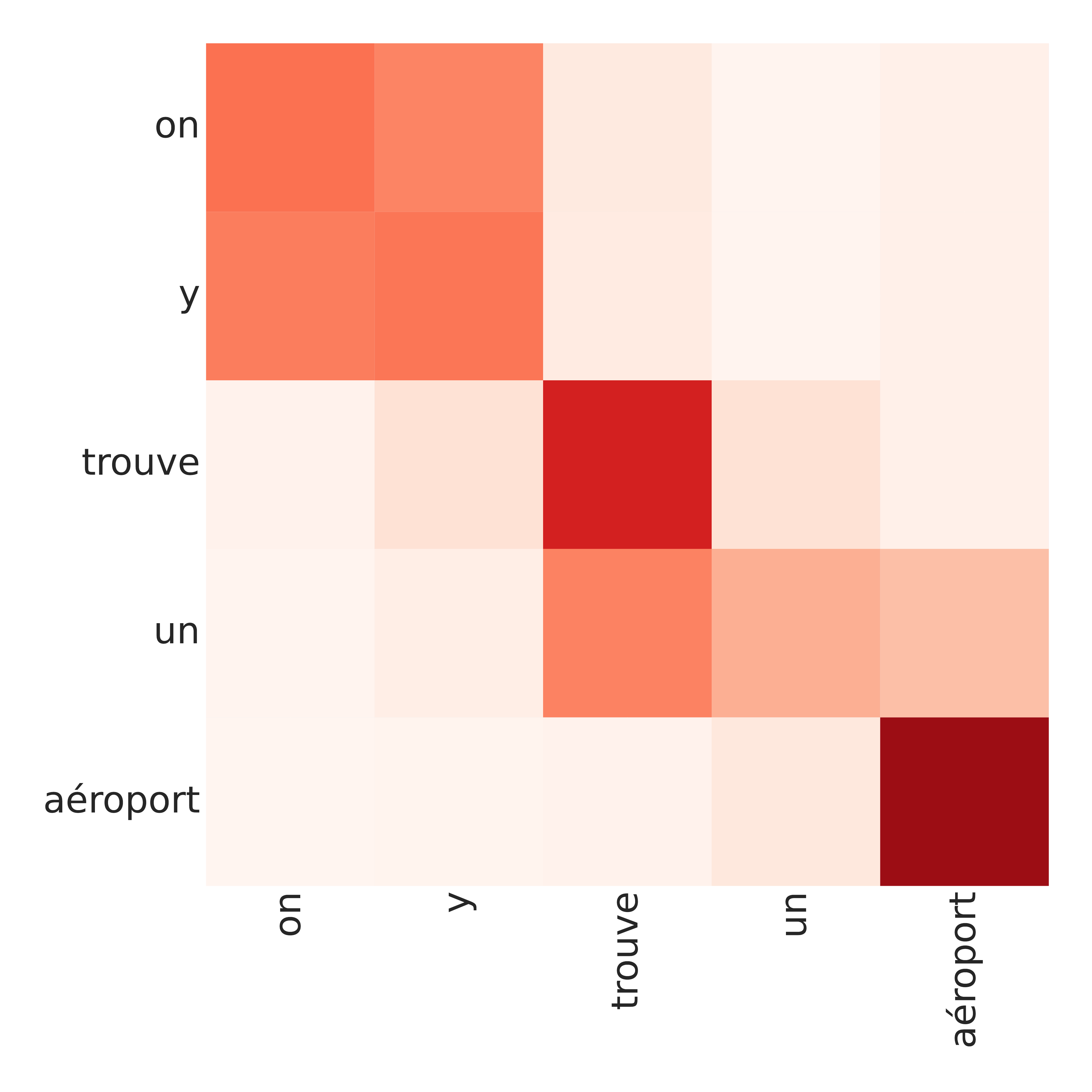}
    \caption{{\bf Within-encoder} Value Zeroing context mixing map for {\bf XLSR-1} at layer 10 for our running example.}
    \label{fig:examplewise_xlsr-1_enc_L10}
\end{figure}

\begin{figure}[th!]
    \centering
    \includegraphics[width=\columnwidth]{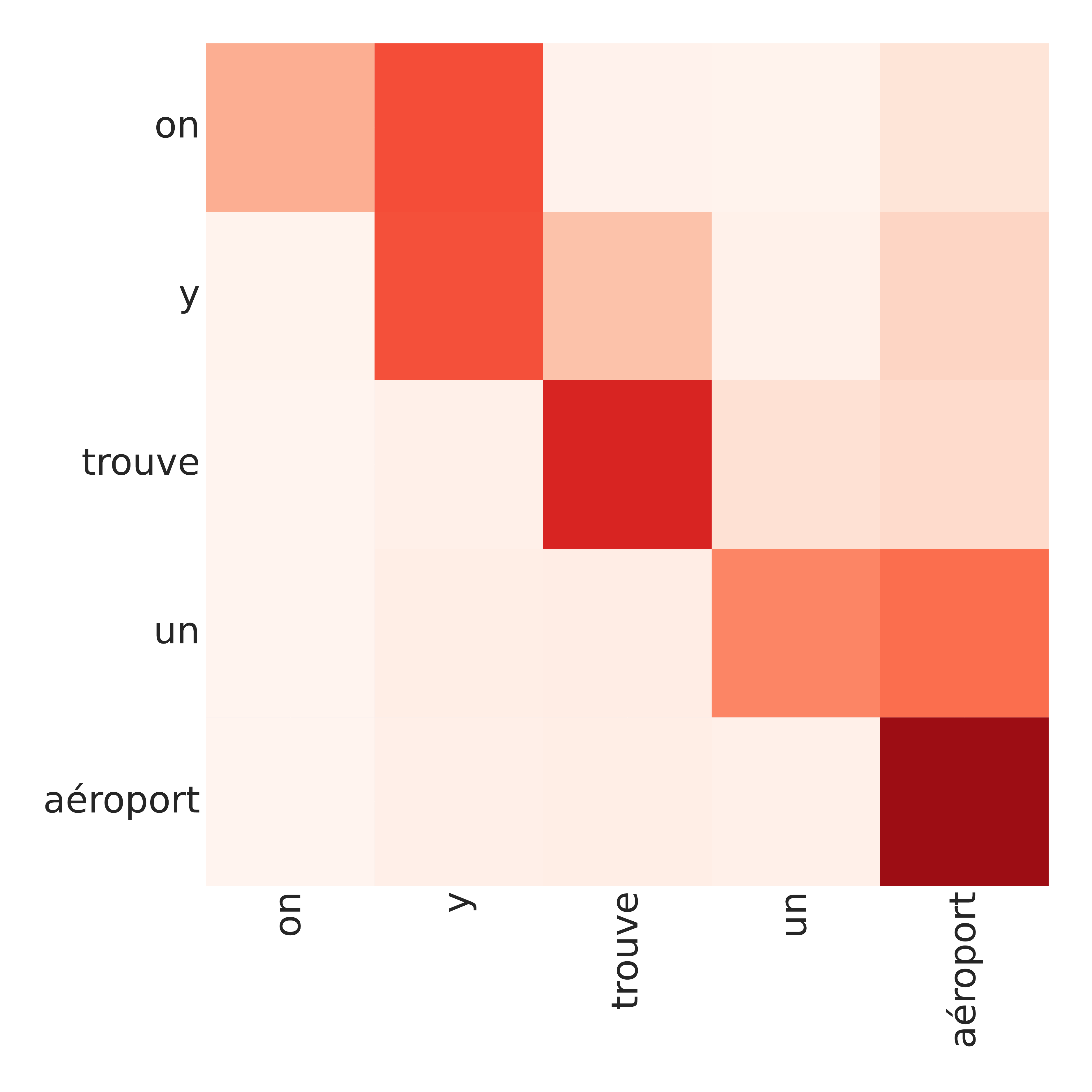}
    \caption{{\bf Within-encoder} Value Zeroing context mixing map for {\bf Whisper-medium} at layer 10 for our running example.}
    \label{fig:examplewise_whisper_enc_L10}
\end{figure}

\begin{figure}[ht!]
    \centering
    \includegraphics[width=\columnwidth]{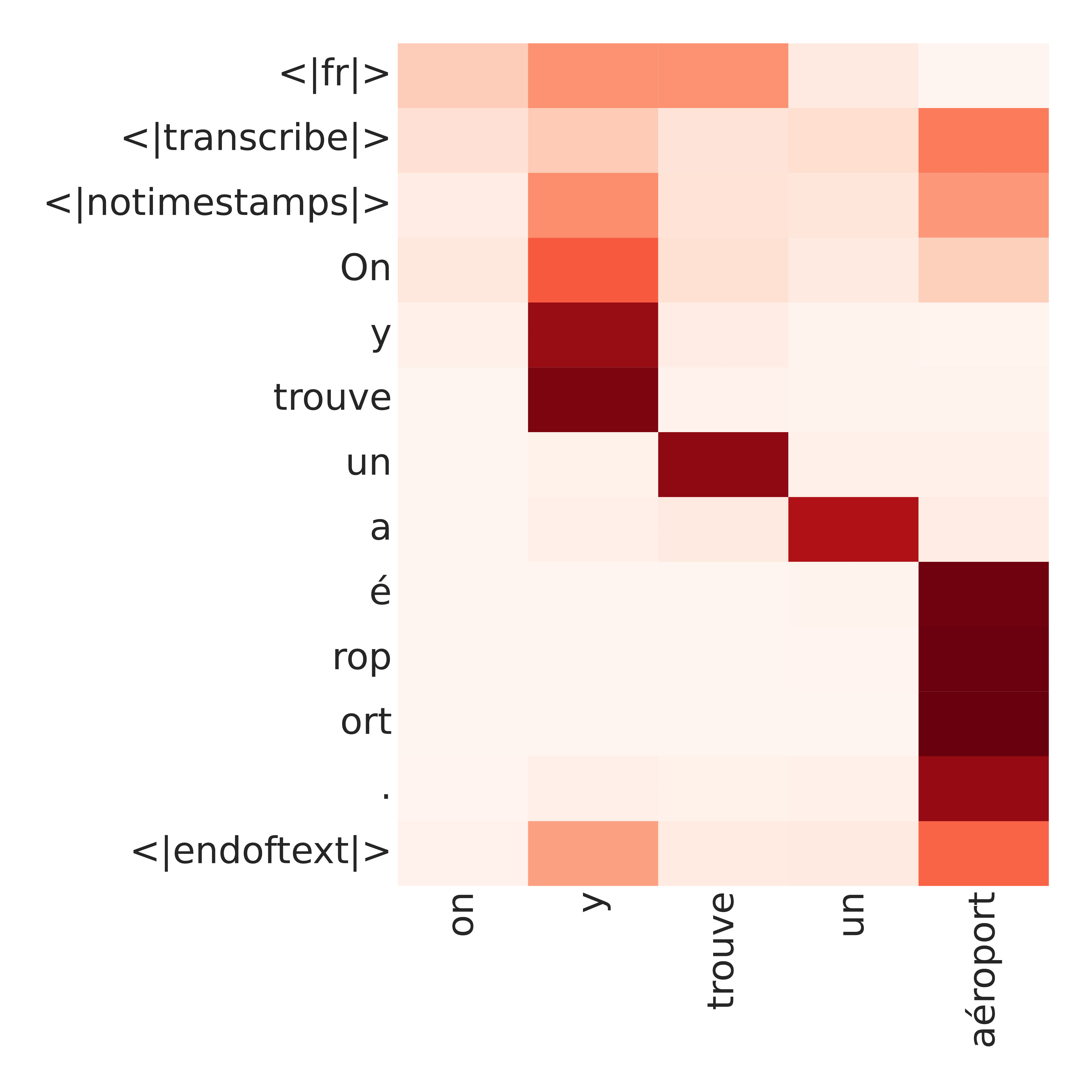}
    \caption{{\bf Cross} Value Zeroing context mixing map for {\bf Whisper-medium} at layer 10 for our running example.}
    \label{fig:examplewise_whisper_cross_L10}
\end{figure}

\end{document}